\documentclass{article}

\usepackage{PRIMEarxiv}

\usepackage[utf8]{inputenc} % allow utf-8 input
\usepackage[T1]{fontenc}    % use 8-bit T1 fonts
\usepackage{hyperref}       % hyperlinks
\usepackage{url}            % simple URL typesetting
\usepackage{booktabs}       % professional-quality tables
\usepackage{multirow}        % multirow cells in tables
\usepackage{amsfonts}       % blackboard math symbols
\usepackage{nicefrac}       % compact symbols for 1/2, etc.
\usepackage{microtype}      % microtypography
\usepackage{lipsum}
\usepackage{graphicx}
\usepackage{subcaption}
\usepackage{xcolor}
\usepackage{amsmath}
\usepackage{amssymb}
\usepackage{amsthm}
\usepackage{bm}
\usepackage{algorithm}
\usepackage{algorithmic}
\usepackage[super]{natbib}

\newtheorem{theorem}{Theorem}

\graphicspath{{media/}{Figures/}}     % organize your images and other figures under media/ folder

\title{Fisher-Informed Parameterwise Aggregation for Federated Learning with Heterogeneous Data}

\author{
  Zhipeng Chang \\
  Department of Mathematics \\
  Penn State University \\
  University Park, PA, USA\\
  \texttt{zfc5231@psu.edu} \\
  \And
  Ting He \\
  Department of Computer Science and Engineering \\
  Penn State University \\
  University Park, PA, USA\\ 
  \texttt{tinghe@psu.edu} \\
  \And
  Wenrui Hao\thanks{*Corresponding author: Wenrui Hao (wxh64@psu.edu)} \\
  Department of Mathematics \\
  Penn State University \\
  University Park, PA, USA\\
  \texttt{wxh64@psu.edu} \\
}

\begin{document}
\maketitle

\begin{abstract}
Federated learning aggregates model updates from distributed clients, but standard first order methods such as FedAvg apply the same scalar weight to all parameters from each client. Under non-IID data, these uniformly weighted updates can be strongly misaligned across clients, causing client drift and degrading the global model. Here we propose Fisher-Informed Parameterwise Aggregation (FIPA), a second-order aggregation method that replaces client-level scalar weights with parameter-specific Fisher Information Matrix (FIM) weights, enabling true parameter-level scaling that captures how each client's data uniquely influences different parameters. With low-rank approximation, FIPA remains communication- and computation-efficient. Across nonlinear function regression, PDE learning, and image classification, FIPA consistently improves over averaging-based aggregation, and can be effectively combined with state-of-the-art client-side optimization algorithms to further improve image classification accuracy. These results highlight the benefits of FIPA for federated learning under heterogeneous data distributions.
\end{abstract}

\vspace{0.5cm}

\section{Introduction}

Federated learning (FL)~\cite{mcmahan2017communication,Kairouz21book} enables collaborative model training across distributed clients without sharing raw data, and has found widespread adoption across diverse domains, including mobile applications (e.g., for Google Keyboard~\cite{googleFL} and Google Assistant~\cite{googleAssistant}), desktop applications (e.g., for Chrome~\cite{FLoC} and Brave~\cite{BraveFL}), multi-centre healthcare~\cite{lee2024international}, cross-hospital prediction~\cite{kuo2025distributed}, edge and IoT intelligence~\cite{li2024introducing}, and security-critical privacy-preserving analytics~\cite{jiang2025towards}. It has also been applied to scientific machine learning to solve partial differential equations (PDEs), including physics-informed neural networks (PINNs) and operator learning with DeepONets~\cite{zhang2025federated,moya2022fed,lu2021learning,zhang2025deeponet,lee2024deep}.

A central challenge in FL is how to update and aggregate model parameters across distributed clients. The most widely used approach is FedAvg~\cite{mcmahan2017communication}, which aggregates local model parameters by taking a data-size–weighted average over all participating clients. While FedAvg performs well when client data are independent and identically distributed (IID), this assumption rarely holds in practical deployments. In real-world settings, client data are typically non-independent and non-identically distributed (non-IID), often exhibiting highly skewed distributions across clients. Such data heterogeneity can cause local model updates to drift away from the global optimum, leading to slow convergence and significant performance degradation in federated learning.

To improve robustness under non-IID data, many methods have been proposed that modify local objectives or training dynamics, such as FedProx~\cite{li2020federated} and drift-correction approaches like SCAFFOLD~\cite{karimireddy2020scaffold}, and further enhance the server update via adaptive ``FedOpt'' variants~\cite{reddi2021adaptive} (e.g., FedAdam/FedYogi/FedAdagrad). In parallel, representation-centric approaches have emerged, notably federated contrastive learning methods such as MOON~\cite{li2021model} and the recent state-of-the-art FedRCL~\cite{fedrcl2023}. Despite their empirical success, these methods typically still adopt client-level aggregation (e.g., FedAvg-style weighted averaging) at the server, where all parameters from each client are scaled by the same scalar weight. This uniform scaling makes it difficult to reflect parameter identifiability from the data—that is, how heterogeneous clients influence different parameters (and directions) unequally—thereby limiting the algorithm’s ability to differentiate updates at the parameter level.

The Fisher Information Matrix (FIM) is widely used to assess parameter identifiability~\cite{miao2011identifiability,komorowski2011sensitivity}. 
Recent work further established a rigorous link between identifiability and the FIM (e.g., via its invertibility/spectrum) and leveraged it for regularization, uncertainty quantification, and experiment design~\cite{wang2025systematic}.
Given that the FIM characterizes how data inform different parameters, recent studies have also incorporated Fisher information into federated optimization to mitigate heterogeneity, including diagonal Fisher-based regularization~\cite{shoham2019overcoming}, Fisher-weighted model fusion~\cite{pmlr-v238-jhunjhunwala24a}, and Newton-type server updates~\cite{pmlr-v162-safaryan22a,siegel2023greedy}. Other approaches have explored duality-based methods with communication acceleration~\cite{park2023dualfl} and algorithmic frameworks for fast federated optimization~\cite{pathak2020fedsplit}.
However, these methods typically rely on diagonal or otherwise overly simplified FIM surrogates that discard parameter identifiability information and fail to capture cross-parameter coupling. As a result, they provide limited expressiveness and cannot achieve truly parameterwise scaling during aggregation.

Inspired by the above gap, we propose \textbf{Fisher-Informed Parameterwise Aggregation (FIPA)}, a server-side aggregation method that explicitly accounts for parameter identifiability encoded in the FIM. In heterogeneous settings, client data induce markedly different FIMs, reflecting variations in parameter identifiability across clients. From a spectral perspective, the eigenvalues and eigenvectors of the FIM indicate which directions in parameter space are well constrained by a client’s data~\cite{wang2025systematic}.

Motivated by this view, we approximate each client’s FIM using its leading-$r$ eigenspace, thereby preserving the most identifiable directions and their relative strengths. This contrasts with diagonal approximations that collapse the FIM and discard cross-parameter coupling and identifiability structure, enabling more expressive and principled parameterwise aggregation.

In each communication round, the server broadcasts the current global model to selected clients. Each client then performs standard local training and, at the received parameter values, computes a rank-$r$ eigendecomposition of its local FIM using only private data. The resulting eigenpairs are uploaded to the server. Using the collected eigenpairs, the server constructs low-rank FIM-based weights and applies them to aggregate client updates in a parameterwise manner. This aggregation is direction-aware, enabling differential scaling across parameters while remaining communication- and computation-efficient.

In this paper, we make the following contributions for FIPA:

\begin{itemize}
\item We improve the computational efficiency of FIPA by approximating the FIM with its dominant identifiable low-rank subspace, together with a warmup–FIPA schedule that controls computation and communication costs while enabling layer-selective fine-tuning for modern deep models.

\item We establish convergence guarantees for FIPA by introducing a three-layer error decomposition framework, using a centralized GN method as a reference.
\item We demonstrate that FIPA consistently improves accuracy and robustness under heterogeneous data distributions across diverse tasks, including nonlinear regression, PDE learning, and image classification, using multiple model families (CNN, ResNet-20, and ResNet-18).
\item We show that FIPA can be seamlessly combined with state-of-the-art client-side optimization algorithms, providing a general server-side mechanism for translating heterogeneous client identifiability into improved federated performance.
\end{itemize}

\section{Results}
\subsection{FIPA Overview and Workflow}

 At communication round $k \in \{1,2,\ldots,K\}$, the server broadcasts the current global model parameters $\bm{\theta}^{(k)}$ to a subset of selected clients. Each client $m \in \{1,2,\ldots,M\}$ computes the top-$r$ eigenpairs of its local FIM using only private data $\mathcal{D}_m$. Specifically, the local FIM, $\mathbf{H}_m^{(k)}$, is approximated by a low-rank decomposition
\begin{equation}\label{eq:lowrank}
\widehat{\mathbf{H}}_m^{(k)} = \mathbf{U}_m^{(k)} \mathbf{\Lambda}_m^{(k)} \big(\mathbf{U}_m^{(k)}\big)^\top,
\end{equation}
where $\mathbf{U}_m^{(k)} \in \mathbb{R}^{p \times r}$ contains the leading eigenvectors and $\mathbf{\Lambda}_m^{(k)} \in \mathbb{R}^{r \times r}$ contains the corresponding eigenvalues.

Each client then performs standard local training initialized at $\bm{\theta}^{(k)}$ and obtains a parameter update $\Delta \bm{\theta}_m^{(k)}$. The client uploads $\Delta \bm{\theta}_m^{(k)}$ together with the eigenpairs $\big(\mathbf{U}_m^{(k)}, \mathbf{\Lambda}_m^{(k)}\big)$ to the server.

\begin{figure*}[t!]
    \centering
    \includegraphics[width=0.8\textwidth,trim=15 15 15 15,clip]{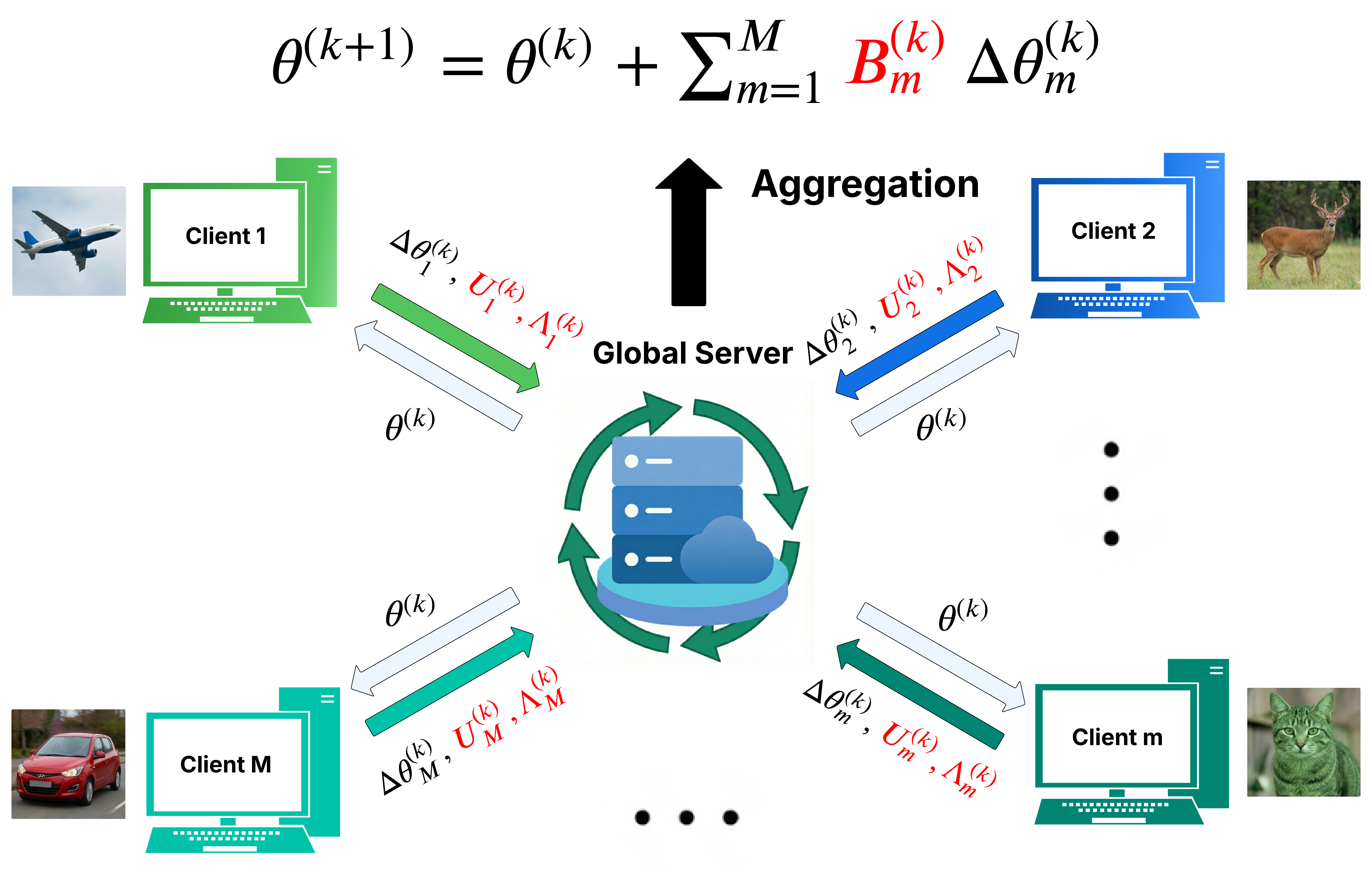}
\caption{Overview of the FIPA workflow and aggregation mechanism. The server aggregates client updates using parameterwise matrix weights $\mathbf{B}_m^{(k)}$ \eqref{eq:bm}. The parameter-server federated learning architecture illustrates the communication flow; heterogeneous client data are shown in different colors.}

    \label{fig:architecture}
    \vspace{-0.3cm}
\end{figure*}

Using the collected eigenpairs, the server constructs low-rank FIM-based aggregation weights
\begin{equation}\label{eq:bm}
    \mathbf{B}_m^{(k)}=\frac{N_m}{N}\Big(\widehat{\mathbf{H}}^{(k)}\Big)^{\dagger}
\Big(\widehat{\mathbf{H}}_m^{(k)}\Big), \quad \widehat{\mathbf{H}}^{(k)}=\sum_{m=1}^{M} \frac{N_m}{N} \widehat{\mathbf{H}}_m^{(k)}.
\end{equation}

where $N_m$ denotes the local data size of client $m$, $N=\sum_m N_m$, and $(\cdot)^{\dagger}$ denotes the Moore--Penrose pseudoinverse. The server then performs parameterwise aggregation to update the global model,
\[
\bm{\theta}^{(k+1)} = \bm{\theta}^{(k)} + \sum_{m=1}^M \mathbf{B}_m^{(k)} \Delta \bm{\theta}_m^{(k)},
\]
which is broadcast to clients in the next communication round.
The overall workflow of FIPA is illustrated in Figure~\ref{fig:architecture}.

For image classification tasks involving large-scale neural networks such as CNNs and ResNets, we employ subspace iteration with Rayleigh--Ritz projection to efficiently compute the leading eigenpairs on the client side, and use thin QR factorization on the server to enable scalable and numerically stable aggregation.

\subsection{Function Fitting}

We study function fitting in both one-dimensional (1D) and two-dimensional (2D) settings, where each client observes training samples only from its own subdomain, resulting in heterogeneous data distributions.

\begin{figure*}[t!]
    \centering
    \includegraphics[width=\textwidth]{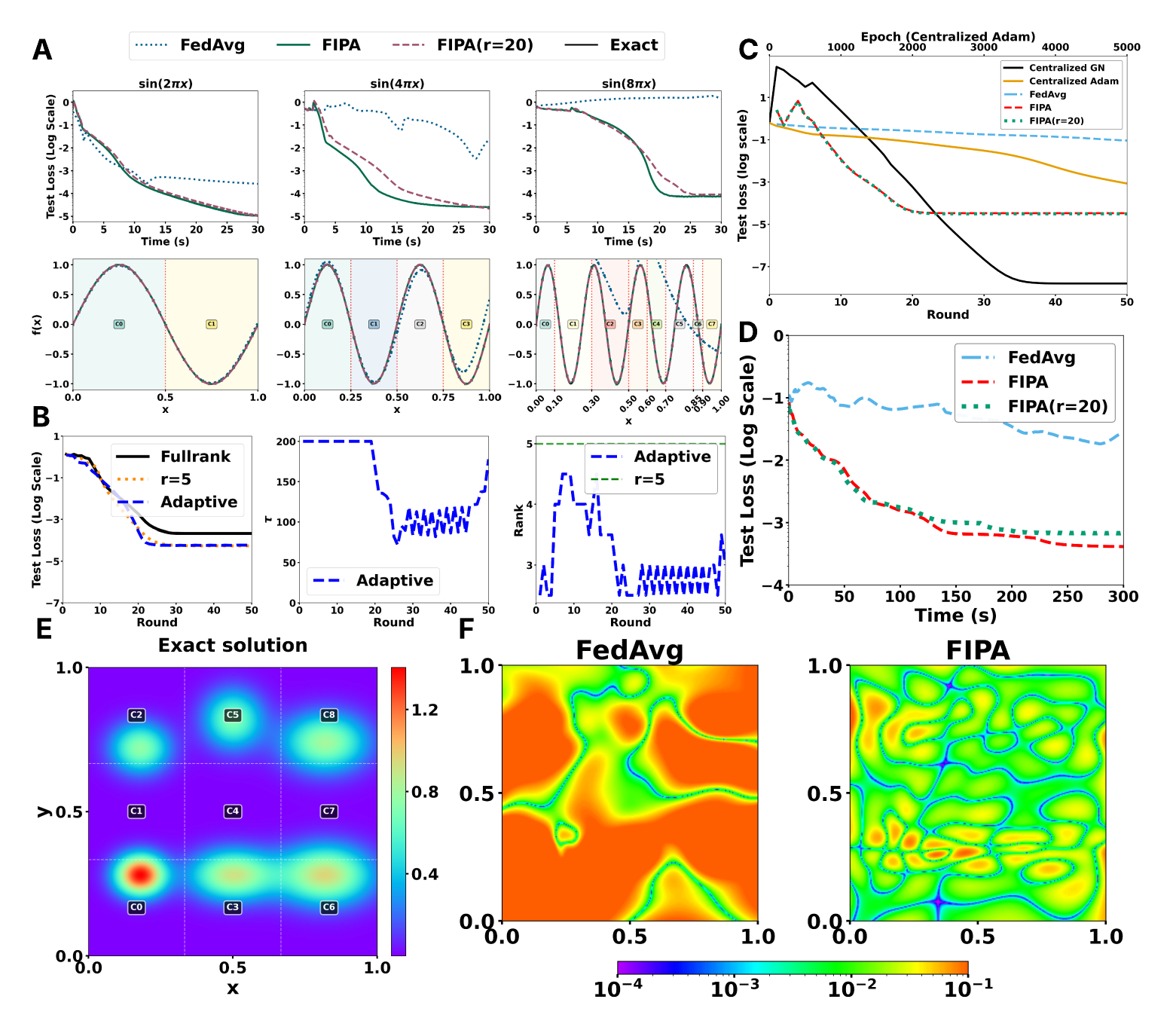}
    \caption{Function fitting results. \textbf{(A)} 1D function fitting: test loss vs. time (top) and fitted function quality (bottom) for target $u(x) = \sin(n\pi x)$ with different client partitions. \textbf{(B)} Adaptive method performance: test loss evolution (left), adaptive local iterations $\tau_m^{(k)}$ (middle), and adaptive rank $r$ (right) for the 2-client $\sin(2\pi x)$ case. \textbf{(C)} Comparison with centralized GN: test loss vs. communication rounds. The black dashed line shows the approximate slope of centralized GN convergence. \textbf{(D--F)} 2D Gaussian-mixture target: test loss vs. time (D), target function and client distribution (E), and pointwise error map (F). }
    \label{fig:d123_fedavg_fedqr}
    \vspace{-0.3cm}
\end{figure*}

\paragraph{1D function fitting.}
We consider target functions of the form $u(x) = \sin(n\pi x)$ with $n \in \{2,4,8\}$. The domain is partitioned into $2$, $4$, and $8$ clients, including non-uniform splits to further increase heterogeneity. The test loss over training rounds and the fitted functions are shown in Fig.~\ref{fig:d123_fedavg_fedqr}A. As the function becomes more oscillatory (larger $n$) or the domain partition becomes more challenging, the performance of FedAvg degrades rapidly. In contrast, FIPA remains stable across all targets and client splits, achieving test losses on the order of $10^{-5}$ and accurately recovering the ground-truth functions. The low-rank variant of FIPA matches the accuracy of the full-rank version while significantly reducing computation and communication costs (Fig.~\ref{fig:d123_fedavg_fedqr}A).

We further evaluate the adaptive low-rank FIM approximation in FIPA on the two-client $\sin(2\pi x)$ case to reduce computational overhead. As shown in Fig.~\ref{fig:d123_fedavg_fedqr}B, FIPA continues to converge as the communication round index $k$ increases, while each client maintains fewer than five eigencomponents throughout training, leading to substantial efficiency gains. The adaptive schedules for $\tau_m^{(k)}$ and $r_m^{(k)}$ are provided in the supplementary.

We also compare FIPA with a centralized Gauss--Newton (GN) baseline~\cite{dennis1996numerical}, where local optimization is performed using Adam~\cite{kingma2014adam}. When Adam is run for sufficiently many steps, the resulting client updates closely approximate a one-step linearized GN update, enabling FIPA to match the convergence rate of centralized GN (Fig.~\ref{fig:d123_fedavg_fedqr}C).

\paragraph{2D function fitting.}
We additionally consider a two-dimensional target function defined as a Gaussian mixture,
\[
u(\mathbf{x}) = \sum_{i=1}^{6} w_i \exp\!\left(-\frac{\|\mathbf{x}-\boldsymbol{\mu}_i\|^2}{2\sigma_i^2}\right), \qquad \mathbf{x} \in [0,1]^2,
\]
where $w_i$ are mixture weights, $\boldsymbol{\mu}_i$ are centers, and $\sigma_i$ are standard deviations. The target function is constructed using six randomly selected Gaussian components to create a multimodal landscape. The domain $[0,1]^2$ is partitioned into a $3\times3$ grid, with each client observing samples only from its corresponding subdomain. As shown in Fig.~\ref{fig:d123_fedavg_fedqr}D--F, FIPA consistently outperforms FedAvg in terms of test loss and error maps across different client partitions.

\subsection{PDE Solving with PINNs}

We consider solving a class of elliptic PDEs of the form
\[
-\Delta u + G(u) = f(\mathbf{x}),
\]
where $G(u)$ denotes a nonlinear reaction term, using the PINNs framework~\cite{raissi2019physics,karniadakis2021physics} with Dirichlet boundary conditions. We study both one-dimensional nonlinear elliptic problems and $d$-dimensional Poisson equations. In our federated setting, boundary conditions are incorporated into each client's local loss function through a penalty term that enforces the boundary constraints. The full local loss function is given in Eq.~\eqref{eq:pde_loss} of the Methods section. The FIM is constructed by concatenating contributions from interior and boundary residuals, as detailed in Eqs.~\eqref{eq:pde_jacobian} and~\eqref{eq:pde_fisher}.

\begin{figure*}[t!]
    \centering
    \includegraphics[width=\textwidth]{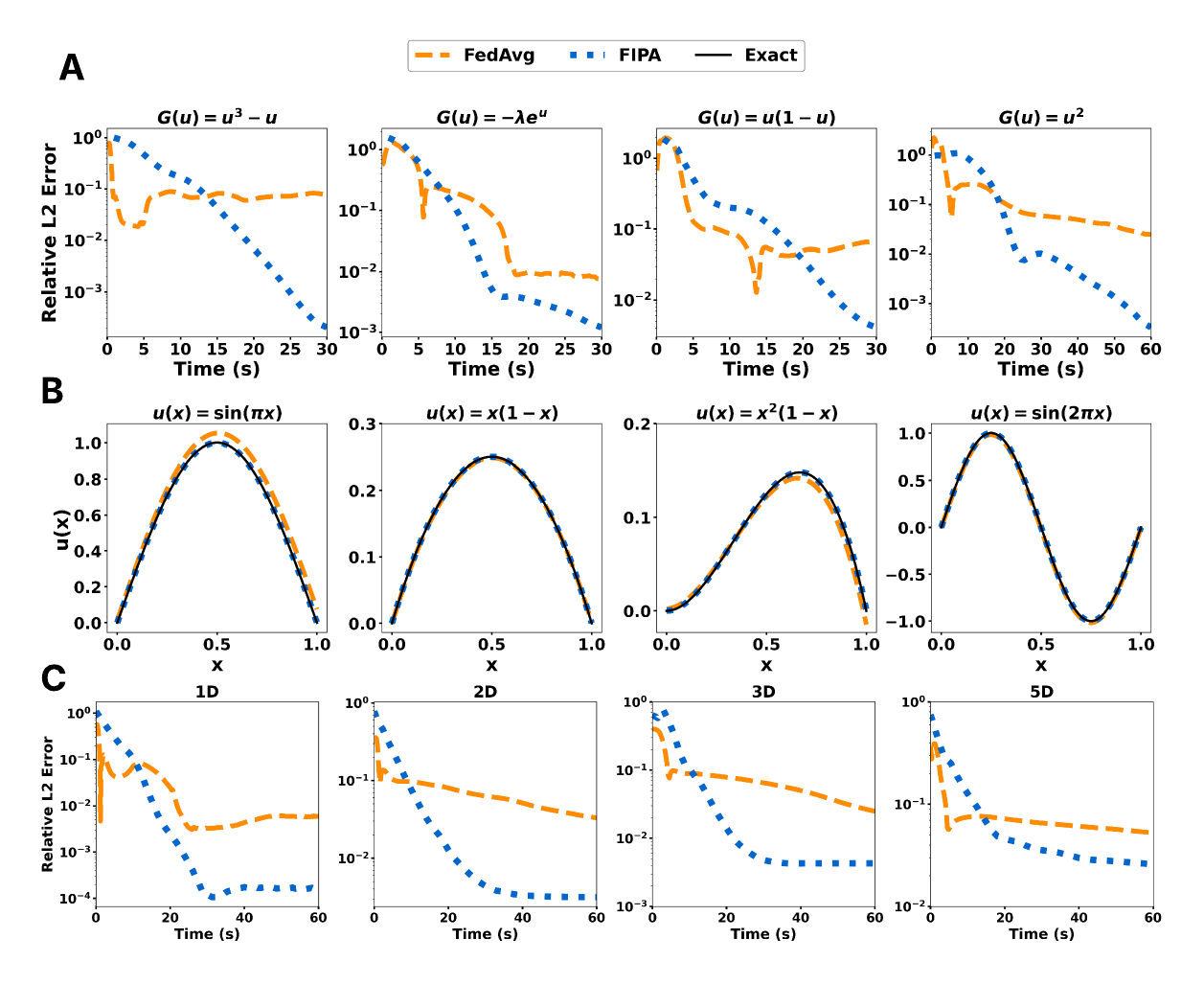}
    \caption{PDE solving results. \textbf{(A)} Relative $L^2$ error vs. time for different 1D nonlinear elliptic PDE problems, from left to right: Allen-Cahn, Bratu, Fisher, Reaction-Diffusion. \textbf{(B)} Solution quality for 1D nonlinear elliptic PDEs. \textbf{(C--E)} $d$-dimensional Poisson equations ($d \in \{1, 2, 3, 5\}$) with two-client split: relative $L^2$ error vs. time.}
    \label{fig:d123_pde_fedavg_fedqr}
    \vspace{-0.3cm}
\end{figure*}

\paragraph{1D nonlinear elliptic equations.}
For one-dimensional problems on $x \in [0,1]$ with Dirichlet boundary conditions, we consider the Allen--Cahn \cite{hao2025stability}, Bratu \cite{bratu}, Fisher \cite{fisher1937wave}, and reaction-diffusion equations \cite{hao2011multiple,lee2025optimal}. Figure~\ref{fig:d123_pde_fedavg_fedqr}A reports the relative $L^2$ error over time for these four nonlinear elliptic PDEs, while Fig.~\ref{fig:d123_pde_fedavg_fedqr}B compares the solution quality obtained by FIPA and FedAvg. Although the evaluation of higher-order derivatives in PINN training increases the computational cost per communication round for FIPA, it consistently achieves lower errors than FedAvg across all problem types. Moreover, as shown in Fig.~\ref{fig:d123_pde_fedavg_fedqr}B, FIPA more accurately captures fine-scale features of the exact solutions.

\paragraph{$d$-dimensional Poisson equations.}
We further evaluate FIPA on $d$-dimensional Poisson equations with $G(u)=0$, where the exact solution is given by
\[
u(\mathbf{x}) = \prod_{i=1}^{d} \sin(\pi x_i),
\qquad
f(\mathbf{x}) = d\pi^2 \prod_{i=1}^{d} \sin(\pi x_i).
\]
To induce data heterogeneity, the domain is split along one coordinate into two clients. Figures~\ref{fig:d123_pde_fedavg_fedqr}C--E show the relative $L^2$ error over time for dimensions $d \in \{1,2,3,5\}$. As the dimension increases, FIPA maintains stable convergence and consistently achieves lower error than FedAvg, demonstrating improved robustness to increasing problem dimensionality.

\subsection{Image Classification on Public Datasets}

We evaluate FIPA on standard image classification benchmarks, including CIFAR-10, CIFAR-100, and Tiny-ImageNet~\cite{krizhevsky2009learning,le2015tiny}, using $M=100$ clients with 5\% participation per round. Non-IID data are simulated via a Dirichlet label distribution~\cite{hsu2019measuring}, controlled by parameter $\alpha$ (smaller $\alpha$ corresponds to stronger heterogeneity). Full training hyperparameters are provided in the supplementary.

In the full-parameter setting, all model parameters are updated. We consider SimpleCNN variants and ResNet-20~\cite{he2016deep}. Training begins with a 1000-round FedAvg warmup, followed by a short FIPA refinement phase of 15 rounds. Experiments on CIFAR-10 are conducted with CNNs of varying sizes (CNN-8k, CNN-23k, CNN-207k; where 8k/23k/207k denote the parameter count) and ResNet-20.

% FIPA's improvements over FedAvg appear immediately after switching from the FedAvg warmup and remain stable across different models and $\alpha$ values.

\begin{figure*}[t!]
    \centering
    \includegraphics[width=\textwidth]{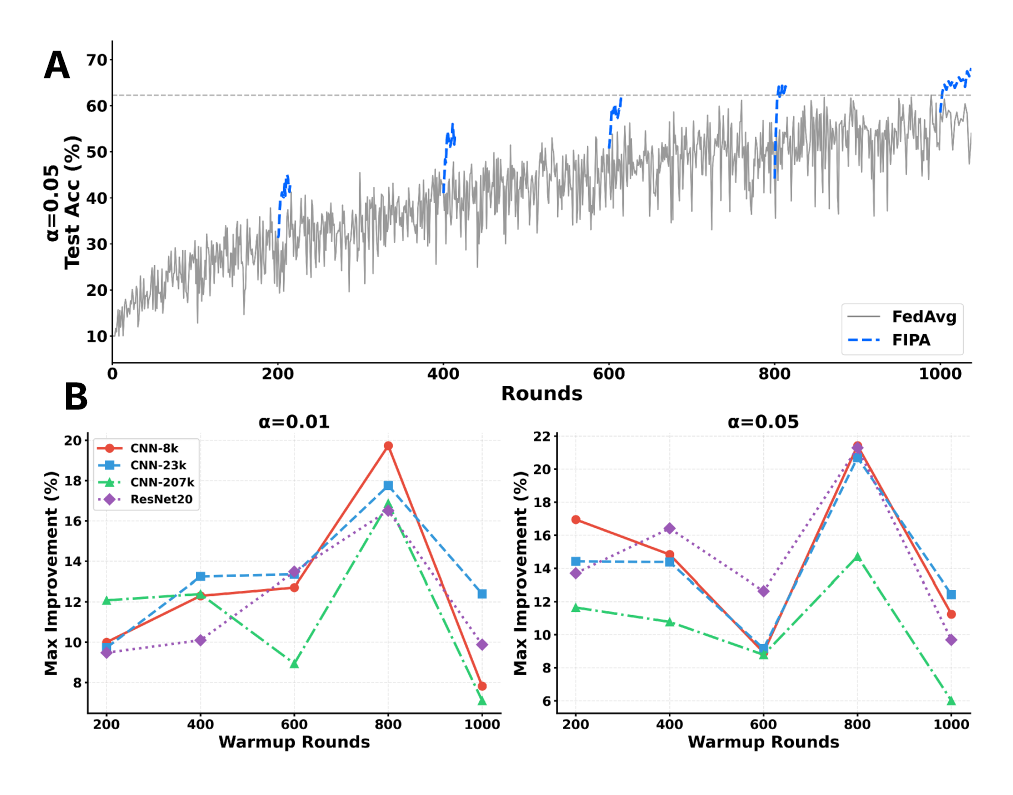}
    \caption{Impact of different warmup rounds on FIPA performance. \textbf{(A)} Test accuracy vs. communication rounds for ResNet-20 under $\alpha=0.05$ when FIPA is initiated after different warmup round counts. \textbf{(B)} Maximum improvement of FIPA over FedAvg within the first 15 rounds after warmup across different network architectures (CNN-8k, CNN-23k, CNN-207k, ResNet-20) and heterogeneity levels ($\alpha \in \{0.01, 0.05\}$).}
    \label{fig:diff_warmupround}
\end{figure*}

Table~\ref{tab:diff_models_results} summarizes top-1 accuracy under different levels of non-IID heterogeneity. FIPA consistently improves accuracy across all heterogeneity levels, with larger gains under stronger non-IID conditions. For example, on CIFAR-10 with $\alpha=0.01$, FIPA increases the maximum accuracy by +9.54\% on CNN-23k (from 51.12\% to 60.66\%) and by +7.35\% on CNN-207k (from 58.45\% to 65.80\%). To investigate the impact of different warmup rounds, we conduct an ablation study where FIPA is applied after varying numbers of FedAvg warmup rounds. As shown in Fig.~\ref{fig:diff_warmupround}, FIPA consistently achieves substantial improvements regardless of the warmup round count.

\begin{table*}[b!]
    \centering
    \caption{Full-parameter training results on CIFAR-10. Training uses FedAvg warmup for 1000 rounds, followed by FIPA and FedAvg for 15 rounds. The table shows top-1 accuracy for four models (CNN-8k, CNN-23k, CNN-207k, ResNet-20) across different non-IID levels. ``Max Accuracy'': best accuracy within 15 refinement rounds. ``Avg Accuracy'': average accuracy over 15 refinement rounds.}
    \label{tab:diff_models_results}
    \small
    \resizebox{\textwidth}{!}{%
    \begin{tabular}{ll*{10}{c}}
        \toprule
        \multirow{2}{*}{\textbf{Metric}} & \multirow{2}{*}{\textbf{Model}} & \multicolumn{2}{c}{\textbf{$\alpha=0.01$}} & \multicolumn{2}{c}{\textbf{$\alpha=0.05$}} & \multicolumn{2}{c}{\textbf{$\alpha=0.1$}} & \multicolumn{2}{c}{\textbf{$\alpha=0.3$}} & \multicolumn{2}{c}{\textbf{$\alpha=0.6$}} \\
        \cmidrule(lr){3-4} \cmidrule(lr){5-6} \cmidrule(lr){7-8} \cmidrule(lr){9-10} \cmidrule(lr){11-12}
        & & \textbf{FedAvg} & \textbf{FIPA} & \textbf{FedAvg} & \textbf{FIPA} & \textbf{FedAvg} & \textbf{FIPA} & \textbf{FedAvg} & \textbf{FIPA} & \textbf{FedAvg} & \textbf{FIPA} \\
        \midrule
        \multirow{4}{*}{Max Acc.} & CNN-8k & 44.09 & \textbf{54.08} & 47.76 & \textbf{56.68} & 54.08 & \textbf{58.74} & 57.37 & \textbf{59.83} & 60.06 & \textbf{62.16} \\
        & CNN-23k & 51.12 & \textbf{60.66} & 57.46 & \textbf{62.89} & 61.94 & \textbf{65.33} & 64.52 & \textbf{67.23} & 67.00 & \textbf{69.29} \\
        & CNN-207k & 58.45 & \textbf{65.80} & 68.07 & \textbf{72.58} & 72.62 & \textbf{75.36} & 76.21 & \textbf{78.40} & 77.75 & \textbf{78.99} \\
        & ResNet-20 & 43.10 & \textbf{51.03} & 61.45 & \textbf{68.19} & 73.49 & \textbf{75.98} & 83.50 & \textbf{85.03} & 85.94 & \textbf{87.09} \\
        \midrule
        \multirow{4}{*}{Avg Acc.} & CNN-8k & 38.80 & \textbf{50.42} & 44.41 & \textbf{53.05} & 48.16 & \textbf{56.33} & 52.63 & \textbf{58.56} & 57.13 & \textbf{60.89} \\
        & CNN-23k & 45.63 & \textbf{56.49} & 52.32 & \textbf{59.93} & 56.11 & \textbf{63.58} & 60.54 & \textbf{65.99} & 65.39 & \textbf{68.77} \\
        & CNN-207k & 53.14 & \textbf{62.53} & 65.14 & \textbf{71.03} & 68.53 & \textbf{73.93} & 73.80 & \textbf{77.39} & 76.31 & \textbf{78.67} \\
        & ResNet-20 & 36.17 & \textbf{47.17} & 56.66 & \textbf{64.80} & 69.25 & \textbf{74.44} & 81.98 & \textbf{84.42} & 85.12 & \textbf{86.70} \\
        \bottomrule
    \end{tabular}%
    }
\end{table*}

To combine FIPA with the current FedRCL method, which uses ResNet-18~\cite{he2016deep} as the base model, we adopt a parameter-efficient fine-tuning strategy inspired by LoRA~\cite{hu2021lora}-style partial updates. For such large-scale models with on the order of $10^7$ parameters, we only update the fully-connected (FC) layers of ResNet-18, which account for 4.4\% of the total parameters, to reduce the computational overhead. This significantly reduces both computation and communication costs.
We warm up with FedRCL for 1000 rounds, and then keep FedRCL as the client-side training rule while switching only the server aggregation from averaging to FIPA to fine-tune the FC layers for another 100 rounds.
Starting from the same warmup checkpoint, we also continue training with FedRCL, FedAvg, and FedProx as baselines for comparison. Table~\ref{tab:fedqr_results} shows the top-1 accuracy results on CIFAR-10, CIFAR-100, and Tiny-ImageNet under both non-IID and IID splits. FIPA improves over FedAvg, FedProx, and FedRCL, with the largest gains when heterogeneity is strong. Under $\alpha=0.05$, FIPA improves Max accuracy by +2.48\% on CIFAR-10 (79.27\% $\rightarrow$ 81.87\%), +1.93\% on CIFAR-100 (54.97\% $\rightarrow$ 57.09\%), and +2.36\% on Tiny-ImageNet (37.38\% $\rightarrow$ 39.86\%) \cite{fedrcl2023}.

\begin{table*}[t!]
    \centering
    \caption{Partial-parameter fine-tuning results on CIFAR-10, CIFAR-100, and Tiny-ImageNet. Training uses FedRCL warmup for 1000 rounds, followed by FIPA, FedProx, FedRCL, and FedAvg for 100 rounds. For FIPA, only the fully-connected (FC) layers are updated using ResNet-18. The table shows top-1 accuracy across different non-IID levels and IID. ``Max'': best accuracy within 100 fine-tuning rounds. ``Avg'': average accuracy over first 100 rounds. Values in parentheses: FIPA's improvement over the best baseline (FedAvg, FedProx, FedRCL).}
    \label{tab:fedqr_results}
    \begin{tabular}{l*{5}{c}}
        \toprule
        \multicolumn{6}{c}{\textbf{CIFAR-10}} \\
        \midrule
        \multicolumn{6}{l}{\textit{Maximum Performance (Best within 100 rounds)}} \\
        \midrule
        \textbf{Method} & \textbf{$\alpha=0.05$} & \textbf{$\alpha=0.1$} & \textbf{$\alpha=0.3$} & \textbf{$\alpha=0.6$} & \textbf{IID} \\
        \midrule
        FedAvg & 79.27\% & 84.42\% & 88.88\% & 89.92\% & 92.04\% \\
        FedProx & 79.39\% & 84.51\% & 89.03\% & 89.98\% & 91.96\% \\
        FedRCL & 79.25\% & 83.98\% & 88.92\% & 89.78\% & 92.03\% \\
        \textbf{FIPA} & \textbf{81.87\%} \textit{(+2.48)} & \textbf{86.17\%} \textit{(+1.66)} & \textbf{89.44\%} \textit{(+0.41)} & \textbf{90.24\%} \textit{(+0.26)} & \textbf{92.30\%} \textit{(+0.26)} \\
        \midrule
        \multicolumn{6}{l}{\textit{Average Performance (Average over first 100 rounds)}} \\
        \midrule
        \textbf{Method} & \textbf{$\alpha=0.05$} & \textbf{$\alpha=0.1$} & \textbf{$\alpha=0.3$} & \textbf{$\alpha=0.6$} & \textbf{IID} \\
        \midrule
        FedAvg & 76.88\% & 82.97\% & 88.09\% & 89.43\% & 91.81\% \\
        FedProx & 77.12\% & 83.06\% & 88.16\% & 89.44\% & 91.79\% \\
        FedRCL & 76.87\% & 82.81\% & 88.11\% & 89.43\% & 91.84\% \\
        \textbf{FIPA} & \textbf{81.53\%} \textit{(+4.41)} & \textbf{85.69\%} \textit{(+2.63)} & \textbf{89.33\%} \textit{(+1.17)} & \textbf{90.14\%} \textit{(+0.71)} & \textbf{92.27\%} \textit{(+0.42)} \\
        \midrule
        \multicolumn{6}{c}{\textbf{CIFAR-100}} \\
        \midrule
        \multicolumn{6}{l}{\textit{Maximum Performance (Best within 100 rounds)}} \\
        \midrule
        \textbf{Method} & \textbf{$\alpha=0.05$} & \textbf{$\alpha=0.1$} & \textbf{$\alpha=0.3$} & \textbf{$\alpha=0.6$} & \textbf{IID} \\
        \midrule
        FedAvg & 54.97\% & 59.62\% & 62.95\% & 64.81\% & 64.78\% \\
        FedProx & 55.16\% & 59.81\% & 63.11\% & 64.92\% & 64.69\% \\
        FedRCL & 55.01\% & 59.79\% & 63.16\% & 65.11\% & 64.78\% \\
        \textbf{FIPA} & \textbf{57.09\%} \textit{(+1.93)} & \textbf{61.41\%} \textit{(+1.60)} & \textbf{64.25\%} \textit{(+1.09)} & \textbf{65.80\%} \textit{(+0.69)} & \textbf{65.18\%} \textit{(+0.40)} \\
        \midrule
        \multicolumn{6}{l}{\textit{Average Performance (Average over first 100 rounds)}} \\
        \midrule
        \textbf{Method} & \textbf{$\alpha=0.05$} & \textbf{$\alpha=0.1$} & \textbf{$\alpha=0.3$} & \textbf{$\alpha=0.6$} & \textbf{IID} \\
        \midrule
        FedAvg & 53.80\% & 58.80\% & 62.54\% & 64.42\% & 64.36\% \\
        FedProx & 53.95\% & 58.96\% & 62.59\% & 64.38\% & 64.30\% \\
        FedRCL & 53.91\% & 58.93\% & 62.68\% & 64.53\% & 64.41\% \\
        \textbf{FIPA} & \textbf{56.32\%} \textit{(+2.37)} & \textbf{60.96\%} \textit{(+2.00)} & \textbf{63.93\%} \textit{(+1.25)} & \textbf{65.51\%} \textit{(+0.98)} & \textbf{65.01\%} \textit{(+0.60)} \\
        \midrule
        \multicolumn{6}{c}{\textbf{Tiny-ImageNet}} \\
        \midrule
        \multicolumn{6}{l}{\textit{Maximum Performance (Best within 100 rounds)}} \\
        \midrule
        \textbf{Method} & \textbf{$\alpha=0.05$} & \textbf{$\alpha=0.1$} & \textbf{$\alpha=0.3$} & \textbf{$\alpha=0.6$} & \textbf{IID} \\
        \midrule
        FedAvg & 37.38\% & 43.68\% & 48.00\% & 49.53\% & 48.94\% \\
        FedProx & 37.47\% & 43.90\% & 48.10\% & 49.53\% & 49.00\% \\
        FedRCL & 37.50\% & 43.55\% & 48.10\% & 49.35\% & 48.72\% \\
        \textbf{FIPA} & \textbf{39.86\%} \textit{(+2.36)} & \textbf{45.41\%} \textit{(+1.51)} & \textbf{48.94\%} \textit{(+0.84)} & \textbf{50.06\%} \textit{(+0.53)} & \textbf{49.03\%} \textit{(+0.03)} \\
        \midrule
        \multicolumn{6}{l}{\textit{Average Performance (Average over first 100 rounds)}} \\
        \midrule
        \textbf{Method} & \textbf{$\alpha=0.05$} & \textbf{$\alpha=0.1$} & \textbf{$\alpha=0.3$} & \textbf{$\alpha=0.6$} & \textbf{IID} \\
        \midrule
        FedAvg & 36.67\% & 42.99\% & 47.66\% & 49.07\% & 48.61\% \\
        FedProx & 36.85\% & 43.23\% & 47.75\% & 49.08\% & 48.68\% \\
        FedRCL & 36.61\% & 43.03\% & 47.62\% & 49.04\% & 48.44\% \\
        \textbf{FIPA} & \textbf{38.74\%} \textit{(+1.89)} & \textbf{44.73\%} \textit{(+1.50)} & \textbf{48.76\%} \textit{(+1.01)} & \textbf{49.86\%} \textit{(+0.77)} & \textbf{48.92\%} \textit{(+0.25)} \\
        \bottomrule
    \end{tabular}
\end{table*}

\section{Discussion}

This work introduces \textbf{Fisher-Informed Parameterwise Aggregation (FIPA)}, a server-side aggregation method for federated learning under heterogeneous data. Standard aggregation methods, such as FedAvg, apply a single scalar weight to the entire client update, scaling all parameters from the same client equally. In contrast, FIPA leverages parameter identifiability and low-rank FIM approximations to quantify how each client's data informs different model parameters, and aggregates updates using parameter-specific matrix weights. Across nonlinear function regression, PDE learning, and image classification, FIPA improves both stability and final performance, with larger gains observed under stronger non-IID splits. For instance, on Tiny-ImageNet, a FedRCL warmup followed by FIPA refinement—updating only the FC head (4.4\% of parameters)—increases top-1 accuracy by 2.36\% over FedRCL.

These results highlight that aggregation should not treat all parameters equally under heterogeneous data. In non-IID settings, clients may provide reliable updates for some parameters but conflicting updates for others, so uniform client-level scaling can amplify drift in poorly aligned directions. By using the FIM to measure local sensitivity, FIPA identifies which parameter directions are well constrained and which are unstable, and reweights updates accordingly. This parameterwise filtering reduces wasted updates and accelerates early-stage convergence while maintaining consistent progress during refinement.

A key practical insight is that FIPA does not require the full FIM to be effective. In our experiments, the useful FIM information consistently concentrates in a small number of dominant features, so the retained rank remains far smaller than the parameter dimension. For compact networks used in function fitting and PDE surrogates, retaining fewer than 20 eigencomponents suffices to closely match full-rank aggregation behavior. As model and task complexity grow, the required rank increases but remains modest relative to the total parameter count. For example, with ResNet-20 (on the order of $10^5$ parameters), keeping fewer than 200 leading features captures the FIM structure relevant for aggregation. For larger models such as ResNet-18 (on the order of $10^7$ parameters), we combine FIPA with parameter-efficient fine-tuning by updating only the FC head, keeping both training and FIM estimation lightweight. In this regime, even a very small rank improves accuracy. 
%Updating only the FC layers on ResNet-20 also reduces communication costs, though it requires longer training, highlighting a trade-off between communication and computation.

Our results motivate a simple two-stage protocol. FIPA is not intended to replace strong first-order federated pipelines from the outset. Instead, a warmup stage—such as FedRCL~\cite{fedrcl2023}—first learns a transferable representation and moves the model into a useful basin, where FIM estimates are more meaningful and less noisy. FIPA is then applied as a short refinement step to redistribute update magnitudes across parameters in a data-aware manner, improving efficiency under heterogeneity. This separation limits overhead while capturing most benefits of Fisher-guided aggregation and aligns naturally with parameter-efficient settings, such as FC-only fine-tuning.

FIPA is complementary to existing client-side approaches for addressing heterogeneity. Methods like FedProx~\cite{li2020federated} introduce proximal regularization to reduce client drift, while representation-based approaches like FedRCL~\cite{fedrcl2023} promote consistent features across clients. These strategies improve non-IID performance, but server aggregation still often uses a single scalar per client. FIPA targets a different part of the pipeline---the aggregation rule itself---using low-rank FIM features to enable parameterwise reweighting. As demonstrated by our FedRCL-based pipeline, we keep the client objective and representation learning unchanged and switch only the server aggregation to FIPA, refining solely how updates are fused at the server.

This study has several limitations. First, FIPA introduces additional computation and communication to estimate and transmit low-rank FIM features. Two trade-offs require careful consideration: the rank truncation, balancing cost versus accuracy, and the fraction of trainable parameters in parameter-efficient fine-tuning, balancing efficiency versus attainable performance. Second, our protocol assumes an idealized setting where each client computes its FIM from the same broadcast model. In practice, stragglers may provide delayed updates, and principled strategies are needed to reuse historical FIM features. Third, while FIPA does not share raw data, FIM-related signals may carry distinct privacy risks compared to standard model deltas, motivating privacy-preserving variants, e.g., via feature perturbation.

Overall, FIPA demonstrates that server aggregation can be improved by moving beyond uniform client-level weighting and instead merging updates with parameter-specific weights derived from compact low-rank FIM features. Combined with a strong client-side warmup and parameter-efficient FC-only fine-tuning, this approach yields consistent gains under heterogeneous data while keeping computation and communication practical. These results position Fisher-informed, parameterwise aggregation as a simple plug-in component for federated learning and motivate future work on robustness to delayed clients and privacy-preserving implementations.

\section{Methods}

\label{sec:methods}

\subsection{Problem setup and notation}
\label{subsec:methods_setup}

We consider a federated learning (FL) system with $M$ clients. Client $m$ holds a local dataset
$\mathcal{D}_m=\{(x_{m,i},y_{m,i})\}_{i=1}^{N_m}$, and $N=\sum_{m=1}^M N_m$. The training goal is to minimize the global empirical risk
\[
F(\bm{\theta})=\sum_{m=1}^{M}\frac{N_m}{N}\,F_m(\bm{\theta}),
\qquad
F_m(\bm{\theta})=\frac{1}{N_m}\sum_{i=1}^{N_m}\ell\!\big(f(x_{m,i};\bm{\theta}),y_{m,i}\big),
\]
where $f(\cdot;\bm{\theta})$ is the model and $\ell(\cdot,\cdot)$ is a generic loss function.

Training proceeds in communication rounds. For $k=0,1,\ldots,K-1$, the server broadcasts $\bm{\theta}^{(k)}$ to selected clients. Each client runs local training initialized at $\bm{\theta}^{(k)}$ and returns an update
$\Delta\bm{\theta}_m^{(k)}=\bm{\theta}_m^{(k)}-\bm{\theta}^{(k)}$.
The server aggregates the received updates and forms $\bm{\theta}^{(k+1)}$. As a reference, the first-order method FedAvg uses
\begin{equation}
\label{eq:fedavg_aggregation}
\bm{\theta}^{(k+1)}=\bm{\theta}^{(k)}+\sum_{m=1}^{M}\frac{N_m}{N}\,\Delta\bm{\theta}_m^{(k)}.
\end{equation}

We use $\bm{\theta}\in\mathbb{R}^{p}$ for model parameters and $f(x;\bm{\theta})\in\mathbb{R}^{C}$ for model outputs. We write $(\cdot)^{\dagger}$ for the Moore--Penrose pseudoinverse. We use $r\ll p$ for the retained rank in low-rank sketches, and denote by
$\mathbf{U}_m^{(k)}\in\mathbb{R}^{p\times r}$ and $\mathbf{\Lambda}_m^{(k)}\in\mathbb{R}^{r\times r}$ the top-$r$ eigenvectors and their associated eigenvalues.
A detailed symbol table is provided in the supplementary.

\paragraph{Fisher information matrix.}
To avoid ambiguity, we state upfront what we mean by ``FIM''.
For a predictive likelihood model $p_{\bm{\theta}}(y\mid x)$, the Fisher information matrix is defined in estimation theory as the expected outer product of the score~\cite{kay1993fssp}:
\[
\mathbf I(\bm{\theta})
:=
\mathbb E_{x}\,\mathbb E_{y\sim p_{\bm{\theta}}(\cdot\mid x)}
\big[\nabla_{\bm{\theta}}\log p_{\bm{\theta}}(y\mid x)\,\nabla_{\bm{\theta}}\log p_{\bm{\theta}}(y\mid x)^{\!\top}\big].
\]
In this work, we use ``FIM'' to refer to the corresponding Fisher curvature used in optimization; for the standard losses considered here, it coincides with the generalized Gauss--Newton (GGN) curvature~\cite{martens2020new}.

\subsection{FIPA aggregation and low-rank approximation}
\label{subsec:methods_hpa}

\textbf{Client side.}
At round $k$, each client $m=1,\ldots,M$ linearizes the model output around $\bm{\theta}^{(k)}$:
\[
f(x_{m,i};\bm{\theta}^{(k)}+\varepsilon)\approx f(x_{m,i};\bm{\theta}^{(k)})+\mathbf{J}_{m,i}^{(k)}\varepsilon,
\qquad 
\mathbf{J}_{m,i}^{(k)}:=\nabla_{\bm{\theta}} f(x_{m,i};\bm{\theta}^{(k)})\in\mathbb{R}^{C\times p}.
\]
Under this linearization, we adopt the {generalized Gauss--Newton} (GGN)~\cite{dennis1996numerical} approximation, which drops
the second-derivative term of the model in the exact Hessian. This yields the client curvature matrix
\begin{equation}\label{eq:hmk}
\mathbf{H}_m^{(k)}:=\frac{1}{N_m}\sum_{i=1}^{N_m}\mathbf{J}_{m,i}^{(k)\top}\,\mathbf{S}_{m,i}^{(k)}\,\mathbf{J}_{m,i}^{(k)}\in\mathbb{R}^{p\times p},
\qquad
\mathbf{S}_{m,i}^{(k)}:=\nabla^2_{\mathbf z\mathbf z}\ell(\mathbf z,y)\big|_{\mathbf z=f(x_{m,i};\bm{\theta}^{(k)})}\in\mathbb{R}^{C\times C}.
\end{equation}
Here, we refer to $\mathbf{H}_m^{(k)}$ as the local FIM (i.e., Fisher/GGN curvature).
For mean squared error (MSE), $\mathbf{S}_{m,i}^{(k)}=\mathbf{I}_C$ and hence $\mathbf{H}_m^{(k)}=\frac{1}{N_m}\sum_{i=1}^{N_m}\mathbf{J}_{m,i}^{(k)\top}\mathbf{J}_{m,i}^{(k)}$.
For softmax cross-entropy, $\mathbf{S}_{m,i}^{(k)}=\mathrm{Diag}(p_{m,i}^{(k)})-p_{m,i}^{(k)}{p_{m,i}^{(k)}}^{\!\top}$ with
$p_{m,i}^{(k)}=\mathrm{softmax}(f(x_{m,i};\bm{\theta}^{(k)}))$, which yields the usual softmax GGN form.

To reduce communication cost and suppress noise, client $m$ transmits the top-$r$ eigenpairs of $\mathbf{H}_m^{(k)}$ and forms
the rank-$r$ approximation
\begin{equation}\label{eq:lowrank_h}
\widehat{\mathbf{H}}_m^{(k)}=\mathbf{U}_m^{(k)}\mathbf{\Lambda}_m^{(k)}\mathbf{U}_m^{(k)\top},
\qquad
\mathbf{U}_m^{(k)}\in\mathbb{R}^{p\times r},\ \mathbf{\Lambda}_m^{(k)}\in\mathbb{R}^{r\times r}.
\end{equation}
Each client starts from the broadcast model $\bm{\theta}^{(k)}$ and performs $\tau$ local optimization steps (e.g., SGD)
to obtain a local iterate $\bm{\theta}_m^{(k,\tau)}$. The uploaded model update is the {total} $\tau$-step change
\[
\Delta\bm{\theta}_m^{(k)}:=\bm{\theta}_m^{(k,\tau)}-\bm{\theta}^{(k)}.
\]
Each client uploads $\big(\Delta\bm{\theta}_m^{(k)},\,\mathbf{U}_m^{(k)},\,\mathbf{\Lambda}_m^{(k)}\big)$ to the server.

\textbf{Server side.}
At the shared linearization point $\bm{\theta}^{(k)}$, the global gradient aggregates as
\[
\mathbf{g}^{(k)}:=\nabla_{\bm{\theta}}F(\bm{\theta}^{(k)})=\sum_{m=1}^{M}\frac{N_m}{N}\,\mathbf{g}_m^{(k)},
\qquad
\mathbf{g}_m^{(k)}:=\nabla_{\bm{\theta}}F_m(\bm{\theta}^{(k)}),
\]
and the (compressed) global curvature aggregates as
\[
\widehat{\mathbf{H}}^{(k)}=\sum_{m=1}^{M}\frac{N_m}{N}\,\widehat{\mathbf{H}}_m^{(k)}.
\]
As a second-order aggregation method, FIPA replaces the scalar weights in FedAvg (Eq.~\eqref{eq:fedavg_aggregation})
with parameterwise matrix weights informed by $\widehat{\mathbf{H}}_m^{(k)}$:
\begin{equation}\label{eq:hpa_update}
\bm{\theta}^{(k+1)}
=
\bm{\theta}^{(k)}+\sum_{m=1}^{M}\mathbf{B}_m^{(k)}\,\Delta\bm{\theta}_m^{(k)},
\qquad
\mathbf{B}_m^{(k)}=\frac{N_m}{N}\big(\widehat{\mathbf{H}}^{(k)}\big)^{\dagger}\widehat{\mathbf{H}}_m^{(k)}.
\end{equation}
This choice is motivated by a consistency property with the {centralized} GN direction:
if the local solver returns the exact minimizer of the quadratic model induced by the same linearization, then
$\Delta\bm{\theta}_m^{(k)}=-(\widehat{\mathbf{H}}_m^{(k)})^{\dagger}\mathbf{g}_m^{(k)}$, and Eq.~\eqref{eq:hpa_update} reduces to
\[
\bm{\theta}^{(k+1)}
=\bm{\theta}^{(k)}-\big(\widehat{\mathbf{H}}^{(k)}\big)^{\dagger}\sum_{m=1}^M\frac{N_m}{N}\mathbf{g}_m^{(k)}
=\bm{\theta}^{(k)}-\big(\widehat{\mathbf{H}}^{(k)}\big)^{\dagger}\mathbf{g}^{(k)},
\]
which matches the centralized GN step (projected to $\mathrm{Range}(\widehat{\mathbf{H}}^{(k)})$ when $\widehat{\mathbf{H}}^{(k)}$ is low-rank).
In practice, $\Delta\bm{\theta}_m^{(k)}$ is produced by $\tau$ local epochs and can be viewed as an inexact solve of this local subproblem.

In practice, the server uses the uploaded top-$r$ eigenpairs ($r\ll p$) to compute the weights directly as
\[
\mathbf{B}_m^{(k)}=\frac{N_m}{N}\Big(\sum_{j=1}^{M}\frac{N_j}{N}\,\mathbf{U}_j^{(k)}\mathbf{\Lambda}_j^{(k)}\mathbf{U}_j^{(k)\top}\Big)^{\dagger}
\Big(\mathbf{U}_m^{(k)}\mathbf{\Lambda}_m^{(k)}\mathbf{U}_m^{(k)\top}\Big).
\]
The detailed algorithm is summarized in Algorithm~\ref{alg:hpa}.

\begin{algorithm}[b]
\caption{Fisher-Informed Parameterwise Aggregation (FIPA)}
\label{alg:hpa}
\begin{algorithmic}[1]
\STATE \textbf{Input:} Global parameters $\bm{\theta}^{(0)}$, number of clients $M$, low-rank dimension $r$, private data $\{\mathcal{D}_m\}_{m=1}^M$ for each client
\FOR{round $k = 0, 1, 2, \ldots$}
    \STATE \textbf{Server:} Broadcast $\bm{\theta}^{(k)}$ to all clients
    \FOR{client $m = 1, 2, \ldots, M$ in parallel}
        \STATE \textbf{Client $m$:} Run $\tau$ local epochs on private data $\mathcal{D}_m$ starting from $\bm{\theta}^{(k)}$ to obtain update $\Delta\bm{\theta}_m^{(k)}$
        \STATE \textbf{Client $m$:} Compute top-$r$ eigenpairs $(\mathbf{U}_m^{(k)}, \mathbf{\Lambda}_m^{(k)})$ of $\mathbf{H}_m^{(k)}$ in Eq.~\eqref{eq:hmk} at $\bm{\theta}^{(k)}$ on $\mathcal{D}_m$
        \STATE \textbf{Client $m$:} Send $(\Delta\bm{\theta}_m^{(k)}, \mathbf{U}_m^{(k)}, \mathbf{\Lambda}_m^{(k)})$ to server
    \ENDFOR
    \STATE \textbf{Server:} Aggregate to obtain $\bm{\theta}^{(k+1)}$ using Eq.~\eqref{eq:hpa_update}
\ENDFOR
\STATE \textbf{Output:} Global parameters $\bm{\theta}^{(k+1)}$
\end{algorithmic}
\end{algorithm}

\textbf{Remark.}
While exact one-step GN updates would theoretically improve consistency with the centralized GN direction, they are computationally expensive and cannot be easily combined with state-of-the-art client-side methods (e.g., FedRCL~\cite{fedrcl2023}). By allowing flexible local training (e.g., SGD with multiple epochs), FIPA maintains compatibility with existing strong baselines while benefiting from Fisher-informed aggregation at the server.

When $C=1$ (scalar outputs), the Jacobian can be written as an $N_m\times p$ matrix $\mathbf{J}_m^{(k)}$ where each row corresponds to a sample.
In this case, Eq.~\eqref{eq:hmk} reduces to $\mathbf{H}_m^{(k)}=\frac{1}{N_m}(\mathbf{J}_m^{(k)})^\top \mathbf{S}_m^{(k)} \mathbf{J}_m^{(k)}$
with a diagonal $\mathbf{S}_m^{(k)}\in\mathbb{R}^{N_m\times N_m}$.
For MSE, $\mathbf{S}_m^{(k)}=\mathbf{I}$ and thus $\mathbf{H}_m^{(k)}=\frac{1}{N_m}(\mathbf{J}_m^{(k)})^\top\mathbf{J}_m^{(k)}$.

For PDE problems with boundary conditions, the local loss becomes
\begin{equation}
\label{eq:pde_loss}
F_m(\bm{\theta})=\frac{1}{2}\|F_{m,\mathrm{int}}(\bm{\theta})\|_{L^2(\Omega_m)}^2+\frac{\beta}{2}\|F_{m,\mathrm{bc}}(\bm{\theta})\|_{L^2(\partial\Omega_m\cap\partial\Omega)}^2
\end{equation}
with penalty parameter $\beta>0$. The Jacobian matrix concatenates interior and boundary terms:
\begin{equation}
\label{eq:pde_jacobian}
\mathbf{J}_m^{(k)}=\begin{bmatrix} \mathbf{J}_{m,\mathrm{int}}^{(k)} \\ \sqrt{\beta}\,\mathbf{J}_{m,\mathrm{bc}}^{(k)} \end{bmatrix},
\end{equation}
and the FIM becomes
\begin{equation}
\label{eq:pde_fisher}
\mathbf{H}_m^{(k)}=(\mathbf{J}_m^{(k)})^\top\mathbf{J}_m^{(k)}=(\mathbf{J}_{m,\mathrm{int}}^{(k)})^\top\mathbf{J}_{m,\mathrm{int}}^{(k)}+\beta\,(\mathbf{J}_{m,\mathrm{bc}}^{(k)})^\top\mathbf{J}_{m,\mathrm{bc}}^{(k)}.
\end{equation}

\subsection{Client-side low-rank FIM sketch for real data}
\label{sec:method_client_sketch}

For large-scale models in real-world data settings, explicitly forming the FIM $\mathbf{H}_m^{(k)}\in\mathbb{R}^{p\times p}$ is computationally prohibitive. We employ subspace iteration with Rayleigh--Ritz projection to extract dominant spectral components efficiently, using only Jacobian-vector products (JVP) and FIM-vector products (FVP) without ever constructing the full FIM.

\textbf{Subspace Iteration.} At client $m$, we identify dominant eigenspaces by iterating on $\mathbf{H}_m^{(k)}$ without forming it explicitly. Starting with random $V^{(0)}\in\mathbb{R}^{p\times L}$ where $L=r+s$ (with oversampling parameter $s$), we compute:
\begin{equation}
    W^{(t)}=\mathbf{H}_m^{(k)} V^{(t-1)}, \quad V^{(t)}=\mathrm{orth}(W^{(t)}), \quad t=1,\dots,q,
\end{equation}
where $\mathrm{orth}(\cdot)$ denotes orthonormalization. The key is that $\mathbf{H}_m^{(k)} V^{(t-1)}$ is computed via FVP accumulated over the full dataset: for each column $v$ of $V^{(t-1)}$, {we compute $\mathbf{H}_m^{(k)} v$ {by enumerating through all $N_m$ local samples as $\mathbf{H}_m^{(k)} v = \frac{1}{N_m}\sum_{i=1}^{N_m} \mathbf{J}_{m,i}^{(k)\top} \mathbf{S}_{m,i}^{(k)} (\mathbf{J}_{m,i}^{(k)} v)$,} using automatic differentiation without storing the $p \times p$ matrix.}
    
\textbf{Rayleigh--Ritz Projection.} We project the FIM onto the identified subspace to obtain a small $L \times L$ matrix. This projection is explicitly computed by enumerating over all local samples $N_m$:
    \begin{equation}
    S_{\mathrm{small}} = {V^{(q)}}^\top \mathbf{H}_m^{(k)} V^{(q)} = \frac{1}{N_m}\sum_{n=1}^{N_m} (\mathbf{J}_n^{(k)} V^{(q)})^\top S_n (\mathbf{J}_n^{(k)} V^{(q)}),
    \end{equation}
where $S_n$ depends on the loss function. Eigendecompose $S_{\mathrm{small}}=\mathbf{W}\mathbf{\Lambda}_{\mathrm{small}}\mathbf{W}^\top$ and retain top-$r$ eigenpairs:
\begin{equation}
    \mathbf{U}_m^{(k)} := V^{(q)} \mathbf{W}_{[:,1:r]}, \quad \mathbf{\Lambda}_m^{(k)}:=\mathrm{diag}(\lambda_1, \cdots, \lambda_r),
\end{equation}
where $\lambda_1 \ge \lambda_2 \ge \cdots \ge \lambda_r$ are the largest eigenvalues of $S_{\mathrm{small}}$. The local FIM is then approximated as $\widehat{\mathbf{H}}_m^{(k)} = \mathbf{U}_m^{(k)} \mathbf{\Lambda}_m^{(k)} \mathbf{U}_m^{(k)\top}$.

\subsection{Server aggregation with QR factorization for real data}
\label{sec:method_server_qr}

For large-scale real-data models, clients transmit only low-rank spectral sketches of the FIM. To apply FIM-preconditioned aggregation efficiently and stably on the server, we solve the resulting system in the sketch subspace of dimension $r_{\mathrm{tot}}=\sum_{m=1}^M r_m\ll p$ using a QR factorization.

With $\widehat{\mathbf{H}}_m^{(k)} = \mathbf{U}_m^{(k)}\mathbf{\Lambda}_m^{(k)} \mathbf{U}_m^{(k)\top}$, the aggregated sketch is
\begin{align}
\widehat{\mathbf{H}}^{(k)} &= \sum_{m=1}^{M}\frac{N_m}{N}\,\widehat{\mathbf{H}}_m^{(k)}
= \mathbf{V}^{(k)}\,\mathbf{\Sigma}^{(k)}\,\mathbf{V}^{(k)\top}, \nonumber \\
\mathbf{V}^{(k)}&=[\mathbf{U}_1^{(k)},\ldots,\mathbf{U}_M^{(k)}], \quad
\mathbf{\Sigma}^{(k)}=\mathrm{blkdiag}\!\Big(\tfrac{N_1}{N}\mathbf{\Lambda}_1^{(k)},\ldots,\tfrac{N_M}{N}\mathbf{\Lambda}_M^{(k)}\Big).
\end{align}
To improve numerical stability when client subspaces overlap, we compute a thin QR factorization $\mathbf{V}^{(k)}=\mathbf{Q}^{(k)}\mathbf{R}^{(k)}$, which yields
\[
\widehat{\mathbf{H}}^{(k)}=\mathbf{Q}^{(k)}\,\mathcal{K}^{(k)}\,\mathbf{Q}^{(k)\top},
\qquad
\mathcal{K}^{(k)} := \mathbf{R}^{(k)}\mathbf{\Sigma}^{(k)}\mathbf{R}^{(k)\top}\in\mathbb{R}^{r_{\mathrm{tot}}\times r_{\mathrm{tot}}}.
\]

Define
\[
\mathbf{b}^{(k)}:=
\sum_{m=1}^{M}\frac{N_m}{N}\,\mathbf{U}_m^{(k)}\mathbf{\Lambda}_m^{(k)}\big(\mathbf{U}_m^{(k)\top} \Delta\bm{\theta}_m^{(k)}\big).
\]
The server update is
\[
\Delta\bm{\theta}^{(k)}
=
\gamma^{(k)}\,(\widehat{\mathbf{H}}^{(k)}+\beta^{(k)} \mathbf{I})^{\dagger}\mathbf{b}^{(k)},
\]
where $\beta^{(k)}>0$ regularizes the pseudoinverse and $\gamma^{(k)}$ is the global step size.
{Note that $\beta^{(k)}$ and $\gamma^{(k)}$ are used only for large-scale image classification; for function fitting and PDE problems, we use Eq.~\eqref{eq:hpa_update} with implicit $\beta=0$ and $\gamma=1$.}
Using the QR form above, the computation reduces to solving
\[
(\mathcal{K}^{(k)}+\beta^{(k)}\mathbf{I})\,\mathbf{z}^{(k)}=\mathbf{Q}^{(k)\top}\mathbf{b}^{(k)},
\]
followed by mapping back to the full space (with a standard orthogonal-complement correction), avoiding any $p\times p$ inversion.

\subsection{Computation and communication}
\label{sec:method_costs}

\paragraph{Computation.}

Relative to FedAvg, the extra computation {(measured in terms of arithmetic time complexity (FLOPs), i.e., the number of scalar multiply–add operations.)} is dominated by building and merging low-rank FIM sketches. On each client, subspace iteration with Rayleigh--Ritz adds a few additional $\mathbf{H}_m^{(k)} v$-type products for a block of $r$ vectors, plus $\mathcal{O}(pr^2)$ orthogonalization and an $\mathcal{O}(r^3)$ eigensolve in the reduced space. On the server, beyond the $\mathcal{O}(p)$ weighted averaging in FedAvg, we perform a thin QR on the stacked bases with cost $\mathcal{O}(p\,r_{\text{tot}}^{2})$ and a reduced-space regularized solve with cost $\mathcal{O}(r_{\text{tot}}^{3})$. In all experiments, $r\ll p$ and the per-round stacked rank remains small, so this overhead is modest compared with local backpropagation.

\paragraph{Communication.}

Each client transmits a local update $\Delta\bm{\theta}_m^{(k)}$ together with a rank-$r$ FIM sketch $(\mathbf{U}_m^{(k)},\mathbf{\Lambda}_m^{(k)})$, requiring $\mathcal{O}(pr)$ scalars per round. With $r$ kept small relative to $p$, the overall communication remains linear in $p$ up to a small constant factor.

\subsection{Two-stage training: warmup and FIPA}
\label{sec:method_warmup_hpa}

The low-rank FIM sketching and QR merging introduce additional computation beyond FedAvg.
To balance efficiency and performance, we adopt a two-stage training pipeline for real data.
In the warmup stage, we run a standard averaging-based federated baseline to obtain a stable initialization.
After warmup, we switch to FIPA and use Fisher-aware aggregation for the remaining rounds.
This design reduces the overall overhead while retaining the benefits of Fisher-informed weighting under heterogeneous data.

\subsection{Parameter-efficient fine-tuning for real data}
\label{sec:method_peft}

For large-scale architectures such as ResNet-18, updating all parameters can be unnecessarily expensive in federated settings.
We therefore follow a parameter-efficient fine-tuning strategy inspired by LoRA~\cite{hu2021lora} and only update a small subset of parameters on each client.
For ResNet-18, we update only the fully-connected (FC) layers, which corresponds to about $4.4\%$ of the model parameters, while the remaining parameters are kept frozen.
This choice substantially reduces both client-side computation and communication, and we show that updating only FC parameters can further improve accuracy over current state-of-the-art methods.

\subsection{Convergence analysis}
\label{subsec:convergence_38}

We analyze FIPA under the least-squares objective
$F(\bm{\theta})=\frac{1}{2N}\|\bm{\xi}(\bm{\theta})\|_2^2$,
where $\bm{\xi}(\bm{\theta})$ denotes the stacked residual vector.
We use a centralized GN~\cite{dennis1996numerical} trajectory as an intermediate reference.
Let $\bm{\theta}^{k,\tau}$ denote the FIPA global model after round $k$ with $\tau$ local epochs,
and let $\Phi_k(\cdot)$ denote one federated FIPA round (local training + server aggregation), so that
$\bm{\theta}^{k+1,\tau}=\Phi_k(\bm{\theta}^{k,\tau})$.

We define the {centralized} damped GN map
\[
T_{\mathrm{GN}}(\bm{\theta}) := \bm{\theta} - \gamma \,\mathbf{H}(\bm{\theta})^{-1}\mathbf{g}(\bm{\theta}),
\]
where $\mathbf{g}(\bm{\theta})=\mathbf{J}(\bm{\theta})^\top \bm{\xi}(\bm{\theta})$ and $\mathbf{H}(\bm{\theta})=\mathbf{J}(\bm{\theta})^\top \mathbf{J}(\bm{\theta})$ with $\mathbf{J}(\bm{\theta}) := \nabla \bm{\xi}(\bm{\theta})$. The reference trajectory is generated as
\begin{equation}
\bm{\theta}_{\mathrm{GN}}^{(0)}=\bm{\theta}^{0,\tau},
\qquad
\bm{\theta}_{\mathrm{GN}}^{(k+1)}=T_{\mathrm{GN}}(\bm{\theta}_{\mathrm{GN}}^{(k)}),
\qquad k=0,1,\dots,K-1.
\label{eq:GN_trajectory}
\end{equation}
We assume the GN system is well-posed: the centralized $\mathbf{H}(\bm{\theta})$ is invertible in a neighborhood of the solution, $\gamma\in(0,1)$ is a fixed damping parameter (consistent with experiments), and we operate in a zero-residual regular regime: there exists a solution $\bm{\theta}^\star$ with $\bm{\xi}(\bm{\theta}^\star)=\mathbf{0}$, the Jacobian $\mathbf{J}$ is Lipschitz continuous near $\bm{\theta}^\star$, and $\mathbf{J}(\bm{\theta}^\star)$ has full column rank (ensuring $\mathbf{H}(\bm{\theta}^\star)\succ 0$).

\textbf{Error decomposition 1: Overall error via centralized GN reference.}
We decompose the overall error into two components:
\begin{equation}
\|\bm{\theta}^{K,\tau}-\bm{\theta}^\star\|
\le
\underbrace{\|\bm{\theta}^{K,\tau}-\bm{\theta}_{\mathrm{GN}}^{(K)}\|}_{\textnormal{federated-to-centralized gap}}
+
\underbrace{\|\bm{\theta}_{\mathrm{GN}}^{(K)}-\bm{\theta}^\star\|}_{\textnormal{centralized reference error}}.
\label{eq:final_split_38}
\end{equation}

\textbf{Error decomposition 2: Federated-to-centralized gap into accumulated and single-round errors.}
Define $e_k:=\|\bm{\theta}^{k,\tau}-\bm{\theta}_{\mathrm{GN}}^{(k)}\|$ and the single-round deviation
\begin{equation}
\delta^{(k)} := \big\|\Phi_k(\bm{\theta}^{k,\tau})-T_{\mathrm{GN}}(\bm{\theta}^{k,\tau})\big\| = \big\|\bm{\theta}^{k+1,\tau}-T_{\mathrm{GN}}(\bm{\theta}^{k,\tau})\big\|.
\label{eq:delta_def_38}
\end{equation}
Using the triangle inequality, we decompose $e_{k+1}$ into two components:
\begin{align}
e_{k+1} &= \|\bm{\theta}^{k+1,\tau}-\bm{\theta}_{\mathrm{GN}}^{(k+1)}\| \nonumber \\
&= \|\Phi_k(\bm{\theta}^{k,\tau})-T_{\mathrm{GN}}(\bm{\theta}_{\mathrm{GN}}^{(k)})\| \nonumber \\
&\le \|\Phi_k(\bm{\theta}^{k,\tau})-T_{\mathrm{GN}}(\bm{\theta}^{k,\tau})\| + \|T_{\mathrm{GN}}(\bm{\theta}^{k,\tau})-T_{\mathrm{GN}}(\bm{\theta}_{\mathrm{GN}}^{(k)})\| \nonumber \\
&= \delta^{(k)} + \|T_{\mathrm{GN}}(\bm{\theta}^{k,\tau})-T_{\mathrm{GN}}(\bm{\theta}_{\mathrm{GN}}^{(k)})\|.
\label{eq:ek_decomp_38}
\end{align}
The first term $\delta^{(k)}$ captures the method difference error (deviation between FIPA and centralized GN updates when starting from the same point), and the second term captures the error from different starting points. By the local contraction property of the Gauss--Newton operator (see the supplementary for detailed proof), $T_{\mathrm{GN}}$ is locally contractive with contraction rate $\rho\in(0,1)$:
\begin{equation}
\|T_{\mathrm{GN}}(\mathbf{x})-T_{\mathrm{GN}}(\mathbf{y})\|\le \rho\|\mathbf{x}-\mathbf{y}\|,
\label{eq:TGn_contract_38}
\end{equation}
which yields $\|T_{\mathrm{GN}}(\bm{\theta}^{k,\tau})-T_{\mathrm{GN}}(\bm{\theta}_{\mathrm{GN}}^{(k)})\|\le \rho e_k$. Combining with~\eqref{eq:ek_decomp_38} gives the recursion:
\begin{equation}
e_{k+1}\le \rho e_k+\delta^{(k)}.
\label{eq:ek_recursion_38}
\end{equation}
With $e_0=0$, unrolling gives
\begin{equation}
e_K\le \sum_{k=0}^{K-1}\rho^{K-1-k}\delta^{(k)}.
\label{eq:accumulation_38}
\end{equation}

\textbf{Error decomposition 3: Single-round error into local training and aggregation errors.}
Introduce an intermediate federated iterate $\widetilde{\bm{\theta}}^{k+1,\tau}$,
obtained from the same starting point $\bm{\theta}^{k,\tau}$ by replacing each client's $\tau$-step local
solver with a single exact local step of the linearized-GN subproblem, while keeping the
same FIPA aggregation on the server. Using the triangle inequality, we decompose $\delta^{(k)}$ into two components:
\begin{align}
\delta^{(k)} &= \big\|\bm{\theta}^{k+1,\tau}-T_{\mathrm{GN}}(\bm{\theta}^{k,\tau})\big\| \nonumber \\
&\le \big\|\bm{\theta}^{k+1,\tau}-\widetilde{\bm{\theta}}^{k+1,\tau}\big\| + \big\|\widetilde{\bm{\theta}}^{k+1,\tau}-T_{\mathrm{GN}}(\bm{\theta}^{k,\tau})\big\| \nonumber \\
&= \delta_k^{(1)} + \delta_k^{(2)},
\label{eq:delta_decomp_38}
\end{align}
where $\delta_k^{(1)} := \|\bm{\theta}^{k+1,\tau}-\widetilde{\bm{\theta}}^{k+1,\tau}\|$ is Error I (local training error from finite local epochs) and $\delta_k^{(2)} := \|\widetilde{\bm{\theta}}^{k+1,\tau}-T_{\mathrm{GN}}(\bm{\theta}^{k,\tau})\|$ is Error II (aggregation error from FIM approximation). Both errors decay as $\mathcal{O}(1/k)$ due to error bounds for local training and low-rank FIM approximation (see the supplementary), hence $\delta^{(k)}\le C/(k+1)$ for some constant $C$.

From $\delta^{(k)}\le C/(k+1)$ and~\eqref{eq:accumulation_38}, we obtain the accumulated gap bound:
\begin{equation}
\|\bm{\theta}^{K,\tau}-\bm{\theta}_{\mathrm{GN}}^{(K)}\|
\le
\frac{C}{1-\rho}\cdot\frac{1}{K}.
\label{eq:gap_rate_38}
\end{equation}
The centralized reference error decays geometrically:
\begin{equation}
\|\bm{\theta}_{\mathrm{GN}}^{(K)}-\bm{\theta}^\star\|\le \rho^K\|\bm{\theta}_{\mathrm{GN}}^{(0)}-\bm{\theta}^\star\|=\rho^K\|\bm{\theta}^{0,\tau}-\bm{\theta}^\star\|,
\label{eq:GN_convergence_38}
\end{equation}
where $\rho\in(0,1)$ is the contraction rate.

\begin{theorem}[End-to-end bound via a centralized damped GN reference]
\label{thm:convergence_38}
Under the zero-residual regular regime (there exists a solution $\bm{\theta}^\star$ with $\bm{\xi}(\bm{\theta}^\star)=\mathbf{0}$, the Jacobian $\mathbf{J}$ is Lipschitz continuous near $\bm{\theta}^\star$, and $\mathbf{J}(\bm{\theta}^\star)$ has full column rank), with fixed damping parameter $\gamma\in(0,1)$ and centralized $\mathbf{H}(\bm{\theta})$ invertible in a neighborhood of the solution, we have
\begin{equation}
\boxed{
\|\bm{\theta}^{K,\tau}-\bm{\theta}^\star\|
\le
\frac{C}{1-\rho}\cdot\frac{1}{K}
+
\rho^K\|\bm{\theta}^{0,\tau}-\bm{\theta}^\star\|,
}
\label{eq:final_bound_38}
\end{equation}
where $\rho\in(0,1)$ is the contraction rate, and $C$ is a constant such that $\delta^{(k)}\le C/(k+1)$ for all $k$.
\end{theorem}

\noindent\textbf{Proof sketch.}

Starting from~\eqref{eq:final_split_38}, we control the federated-to-centralized gap through the
single-round deviation~\eqref{eq:delta_def_38}. The contraction~\eqref{eq:TGn_contract_38} gives the
recursion~\eqref{eq:ek_recursion_38} and the weighted accumulation~\eqref{eq:accumulation_38}. The
$\mathcal{O}(1/k)$ decay of Error I/II implies $\delta^{(k)}=\mathcal{O}(1/k)$, which yields the
$\mathcal{O}(1/K)$ gap bound. The reference term follows from the geometric decay $\rho^K$ for the centralized damped GN trajectory, while the gap term follows from the recursion $e_{k+1}\le \rho e_k+\delta^{(k)}$ and $\delta^{(k)}=\mathcal{O}(1/k)$.

\section*{Reporting summary}

Further information on research design is available in the Nature Portfolio Reporting Summary linked to this article.

\section*{Data availability}

The datasets used in this study are publicly available: CIFAR-10 and CIFAR-100~\cite{krizhevsky2009learning} are available from \url{https://www.cs.toronto.edu/~kriz/cifar.html}, and Tiny-ImageNet is available from \url{https://www.kaggle.com/c/tiny-imagenet}.

\section*{Code availability}

Code and scripts to reproduce the experiments are available at https://github.com/changzhipeng1-prog/FIPA.git, including configuration files and seeds.

\bibliographystyle{naturemag}

\bibliography{references}

\section*{Acknowledgments}
This research was supported by National Institute of General Medical Sciences through grant 1R35GM146894 (ZC and WH) and National Science Foundation under award CNS-2106294 (TH).

\section*{Author contributions}

W.H. conceived the study. W.H. and Z.C. developed the methodology and designed the experiments. Z.C. implemented the code and performed the experiments. Z.C. analyzed the results and prepared the figures. Z.C. wrote the initial draft. W.H., Z.C. and T.H. reviewed and edited the manuscript and approved the final version.

\section*{Competing interests}

The authors declare no competing interests.

\section*{Additional information}

Correspondence and requests for materials should be addressed to Wenrui Hao (wxh64@psu.edu).
\end{document}